# Improved Neural Modeling of Real-World Systems Using Genetic Algorithm Based Variable Selection

Donald A. Sofge & David L. Elliott
NeuroDyne, Inc.
One Kendall Square
Cambridge, MA 02139, USA
sofge@ai.mit.edu, delliott@isr.umd.edu

**Abstract**
Neural network models of real-world systems, such as industrial processes, made from sensor data must often rely on incomplete data. System states may not all be known, sensor data may be biased or noisy, and it is not often known which sensor data may be useful for predictive modelling. Genetic algorithms may be used to help to address this problem by determining the near optimal subset of sensor variables most appropriate to produce good models. This paper describes the use of genetic search to optimize variable selection to determine inputs into the neural network model. We discuss genetic algorithm implementation issues including data representation types and genetic operators such as crossover and mutation. We present the use of this technique for neural network modelling of a typical industrial application, a liquid fed ceramic melter, and detail the results of the genetic search to optimize the neural network model for this application.

**Introduction**
When modeling a complex system (such as a chemical reactor), it is not generally known *a priori* which system states are necessary to develop a good model, or which states are observable based upon available sensor technology (although it is often known that many system states are *not* observable). In addition, there is a greater problem in identifying useful data. Complex dynamic systems such as the chemical reactor may be instrumented with tens, hundreds or even thousands of sensors. The problem with so much sensor information is that most of it will be irrelevant. Worse still, unfiltered incorporation of irrelevant data will adulterate a model, eroding its predictive capabilities.

A key data pretreatment problem is sensor redundancy. It is well known that smaller models are often better models [Sofge92]. This translates to fewer inputs and fewer hidden layer nodes. While it may be nice to have highly redundant data from a large number of sensors, in reality we may only need a few key sensors in order to produce a good model. The problem is in determining which few sensors to choose, and ignoring most of the remaining sensors. This is confounded by the fact that due to differing sensor response characteristics and noise, in the aggregate there is a considerable amount of noise and bias in the data.

In the example given in this paper, modelling of a liquid fed ceramic melter (LFCM) process is undertaken in order to predict the surface level. The melt chamber is instrumented with 20 thermocouple sensors placed at different sites within the chamber. Each sensor may have a slightly different characteristic response curve due to differences in manufacturing, usage history, etc. Each sensor also is susceptible to some level of noise. We take a time history of data from all 20 sensors and store it in our database, and then use this database to train a neural network model.

Some sensors, such as those near the surface in the reactor vessel, may offer fairly high-variance data throughout the process, but be largely irrelevant to accurately predicting final product quality. We would like to select a near- optimal set of sensor variables in order to train a neural network model with the greatest predictive accuracy.

**Variable Selection Using Genetic Algorithm**
A genetic algorithm (GA) is fundamentally a search method which is used to optimize a complex system which is too large to fully explore or to locate a true optimal solution. The GA search procedure is inspired by rules of natural selection in Darwinian evolution which suggest that only the fittest members of a group will survive, to then recombine genetically with other fit members to yield even fitter members, thereby passing their successful characteristics on to the next generation [Holland75]. Less competitive members of the group are discarded or die off and are not recombined, and thus the characteristics that they carry are not propagated. Thus a population "evolves", with successive generations replacing older ones and more successful members replacing less successful ones. Each member of the population, called a "chromosome", is represented by a string of "genes", which are encoded characteristics to be optimized.

The genes need to be defined for a given application such that finding a better or more optimal set of genes means finding a better solution to the problem. A GA may perform variable selection if each gene in a chromosome represents an available sensor variable. Fitness is judged for each chromosome by determining how good the models are (accuracy, robustness) generated by that combination of variables. An initial population of chromosomes is generated by choosing a string length (# of genes) and randomly assigning a variable to each gene. The GA search is then set in motion and the chromosomes compete, reproduce, and die off as they are replaced by more fit chromosomes. It is usually desirable to maintain a fixed-size population in order to make sure that the fitter chromosomes quickly replace the less fit ones. An occasional mutation is introduced to make sure that certain genes (variables) which may be really useful aren't quickly eliminated (possibly because they are randomly combined with really noisy variables early on) and then never incorporated again. This is referred to as a population in danger due to lack of genetic variation, and to avoid this situation a mutation rate is predetermined and mutated chromosomes are introduced into the population at regular intervals during GA search. As these parameters are application dependent, it is not possible to know beforehand which values will work best. The GA process is automated with automatic gene sequence selection, model building and discarding, and evaluation of accuracy and robustness of the models (scoring). Successive generations will inherit the best characteristics from the previous generation, while eliminating the less valuable characteristics.

**GA Representation**
Genetic algorithms are often thought of, discussed and implemented using binary strings, or bit strings. Each bit represents a "gene" expression. If the bit is turned on, then the gene corresponding to that bit can be said to be "expressed". While this representation works fine for most purposes, it is not necessary to use binary representations to implement genetic algorithms. In this project each chromosome (representing a subset of selected input variables for the neural network model) is expressed as a vector of integers, with each integer representing a particular gene. For certain GA operators, however, such as the crossover operator (discussed below), this form is translated into an equivalent binary form in order to facilitate simpler computation using expressions of binary logical operators. The resulting offspring is then translated back to its equivalent vector form. In general it is best to use whatever representation format for storing and manipulating the chromosomes is most appropriate for the application domain.

**GA Operators**
Genetic algorithms require the use of special operators in order to simulate the evolutionary processes which they emulate. The most important are the crossover and mutation operators. The crossover operator takes two parent chromosomes (in this application, each parent chromosome represents a group of input variables used to build a neural network model), and combines them to produce an offspring. The most common form of crossover operator in GA literature is known as uniform crossover [Spears91]. In uniform crossover, if a specific gene is turned on in both parents, then it will be turned on in the offspring. If a gene is turned on in only one of the parents, then it may be turned on (with a predetermined probability, usually 0.5) in the offspring. Uniform crossover was used in this project as well.

The mutation operator is applied independently but immediately following the crossover operator. A mutation is a random addition or deletion of a gene in a chromosome, and is governed by a preset mutation rate (usually quite low, e.g. 0.001).

A technique not as commonly used in GA literature, but developed for this application, is a survival rate, which determines what percentage of the population (the fittest members) will survive to be continued into the next generation. Many early applications of GAs assumed that all chromosomes of a generation would be replaced by their offspring. However, this was found to often eliminate the fittest chromosomes and interfere with the search. An approach called "elitism" or "superiority" [deGaris93] was employed to keep the fittest chromosome from a previous generation. In this project, a percentage (e.g., top 20%) of the chromosomes from the previous generation were carried on to the next generation. These top performers were used to generate the offspring for the following generation.

Another feature employed in this work was to guarantee that when a new offspring is generated it does not duplicate any current chromosome in the population. A graveyard is used to store old chromosomes which represent models which have been built, tested, and then discarded. Each new offspring is compared with chromosomes in the graveyard to make sure that it hasn't been tested before in a previous generation. Since we assume that all of the neural network models use the same superset of data (same output data, input data includes sensor streams for all possible input variables), then the process of choosing variables for a particular model is deterministic, so there is never a need to retest a chromosome once its corresponding model has been built, tested and scored. This promotes better crossover by preventing the generation of chromosomes which are already represented or have been generated and tested in prior generations. Chromosomes which are carried from one generation to the next are stored along with their scores, but are not retested since this would serve no purpose.

**Simulation Results**

The process being modelled in this effort is a liquid fed ceramic melter (LFCM). The LFCM is instrumented with 20 thermocouples distributed throughout the melt chamber which provide temperature feedback during the process. Data from these 20 sensors, 200 samples of each taken at a specific interval of time, is collected along with a measurement of level in the melt chamber (see Figure 1). This data is used to train the neural network models.

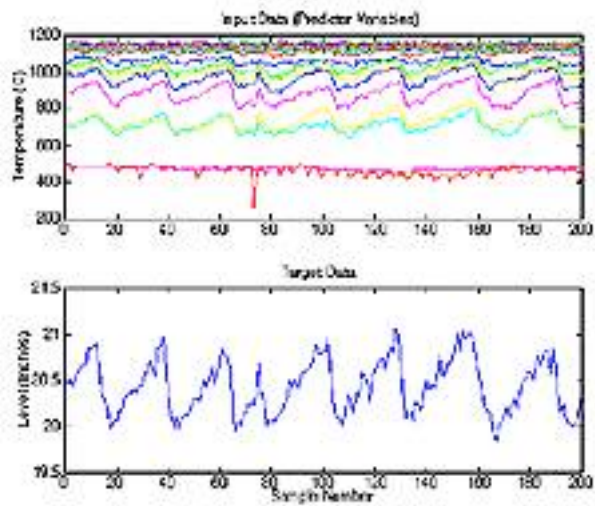

Figure 1. Training Data for LFCM Process

As shown in the top part of Figure 1, the thermocouple readings (20 readings overlaid onto the same plot) are quite noisy. In some there is no apparent correlation between the sensor readings and the level measurement shown in the bottom part of the figure. Also, there is considerable variability in the response of various sensors (which may be due to each sensor's location in the chamber, or due to the response characteristics of the sensor itself, or both).

In order to examine every possible grouping of input variables (not including permutations, only combinations) to find the optimal subset of input parameters for modelling the level in the LFCM, it is useful to think of a bit string of length 20. A bit turned on would indicate that that variable was included in the solution. Excluding the all zeros case (where no inputs are used), there are $2^{20}-1$ or 1,048,575 unique models which can be formed using these inputs. It is clearly unreasonable to try to build, train and test this many neural network models. Since we don't know a priori which inputs will used, we need a procedure for finding a near optimal subset. The GA provides the solution.

Each combination of input variables, which can be thought of as a 20-bit string, is a chromosome. In the GA procedure we build a fixed size population of chromosomes, which are each evaluated and scored according to a fitness function. In this application the models were trained on the training data (200 exemplars), and then tested using an independent cross-validation dataset not used for training. The cross-validation data consisted of 200 exemplars. The neural networks all used the same number of hidden-layer and output nodes, and the same non-linear activation function. The neural networks were multilayer perceptrons trained using the Levenberg-Marquardt algorithm (which generally converges more quickly than standard back-propagation). Each network was allowed to train to completion. The score for each model, or chromosome, is simply the sum-squared-error (SSE) obtained from applying each network to the cross-validation dataset. The goal of the genetic search then is to find the model with the minimum total SSE on the cross-validation dataset.

The population was initialized using a combination of ordered and random selection of chromosomes. First, all of the single variable subsets were included, and the solution utilizing all of the input variables was included. In between these extremes, the database was populated with a roughly even (by the number of genes expressed) distribution of chromosomes, though whether a particular gene was expressed in a particular chromosome was determined by a random number generator.

Various runs were made using population ranges from 30 to 100 chromosomes. The survival rate was varied between 20% and 50%, and various mutation rates were tried. Crossover was achieved by random selection of the fittest chromosomes from the previous generation. As noted in GA literature [Goldberg89, Davis91] use of a uniqueness operator as employed in this effort allows higher crossover and mutation rates, and enables more rapid convergence of the genetic search procedure. A mutation rate of 0.1 (usually the mutation rate is closer to 0.001) was found to work quite well in this instance.

After several runs of 20 to 30 generations (1000-2000 models built and tested for each run) the genetic search returned the same result each time as the best solution, despite use of different randomly generated populations, different population sizes, and different GA operator settings. Out of 20 input variables, numbered 1 through 20, it found that the best model resulted from selection of 11 of these variables: 1-2-3-4-5-8-9-14-16-18-19. The genetic search procedure excluded the 9 variables 6-7-10-11-12-13-15-17-20. Whether this is the optimal solution is an open question without exhaustively testing all 1,048,575 possible models and comparing their scores. The final solution on the cross-validation dataset is shown in Figure 2.

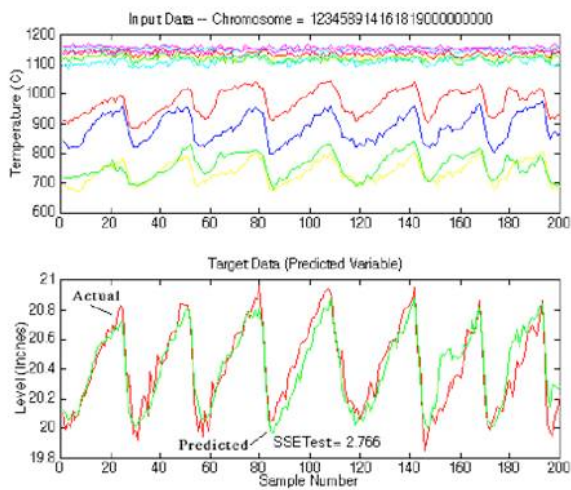

Figure 2. Final Solution: Model Tested on Cross-Validation Data

**Conclusions**

This paper presents a solution to the problem of trying to build neural network models of real-world systems, such as chemical and industrial processes, using data from numerous sensors where sensor data may be noisy, biased, corrupted, or even irrelevant to the parameter(s) being modelled. The use of genetic search makes it possible to find a near-optimal subset of variables for use in model-building under conditions where the data may make this quite difficult. The technique of GA based variable selection may be applied to numerous application areas where models (neural network or other) are required, the selection of input variables is not always clear, and the data may be noisy. A typical example of this would be in financial forecasting.